# Adaptive Intelligent Cooperative Spectrum Sensing In Cognitive Radio

Dilip S. Aldar

**Abstract** –*Radio Spectrum is most precious and scarce resource and must be utilized efficiently and effectively. Cognitive radio is the promising solutions for the optimum utilization of the scared natural resource. The spectrum owned by the primary user should be shared among the secondary user, but primary user should not be interfered by the secondary user. In order to utilize the primary user spectrum, secondary user must detect accurately, the existence of primary in the band of interest. In cooperative spectrum sensing, the channel between the secondary users and the cognitive radio base station is non stationary and causes interference in the decision in decision fusion and in information in information due to multipath fading. In this paper neural network based cooperative spectrum sensing method is proposed, the performance of proposed method is evaluated and observed that, the neural network based scheme performance improve significantly over the AND,OR and Majority rule.*

## I. Introduction

The spectrum is natural valuable resource, and in the recent year use of the spectrum increased exponentially due to emerge of the new technology and applications, which results the scarcity of spectrum. The spectrum must be utilized efficiently, the cognitive radio is the one of prospective solutions for efficient utilization of the radio spectrum at a specific time and location. The primary user is who have a higher priority or legacy rights for the use of a specific band of spectrum. Whereas, secondary users with lower priority, exploit the spectrum in without interference to primary users. Therefore, secondary users must have the cognitive capabilities to sense the spectrum reliably, to check it is being used by a primary user, and should exploit the unused part of the spectrum [1].

The spectrum is allocated to the licensed user is not completely used all the time and in all the places [2]. The availability of the spectrum at a specific time and location should be detected efficiently to provide highly reliable communication whenever and wherever needed [3]. To detect the existence of free spectrum, the efficient spectrum sensing technique must be employed. If the secondary user is in the deep fading may not be able to make appropriate decisions about the existence of primary (licensed) user is the problem of hidden terminal. The cooperative spectrum sensing is promising method to deal with hidden node problem, and to acquaint the white spaces to secondary (cognitive radio) users. Fig. 1 illustrates the cognitive radio system with cooperative spectrum sensing, where secondary users sense the primary and communicate their decision to the cognitive radio base station is decision fusion, and the secondary user transmits the data to the cognitive radio base station is information fusion.

As the spectrum sensing is an important task of the cognitive capabilities, the several aspects of spectrum sensing are discussed in [1] such as multidimensional spectrum sensing, external sensing algorithms, spectrum sensing challenges in the hardware perspective etc.

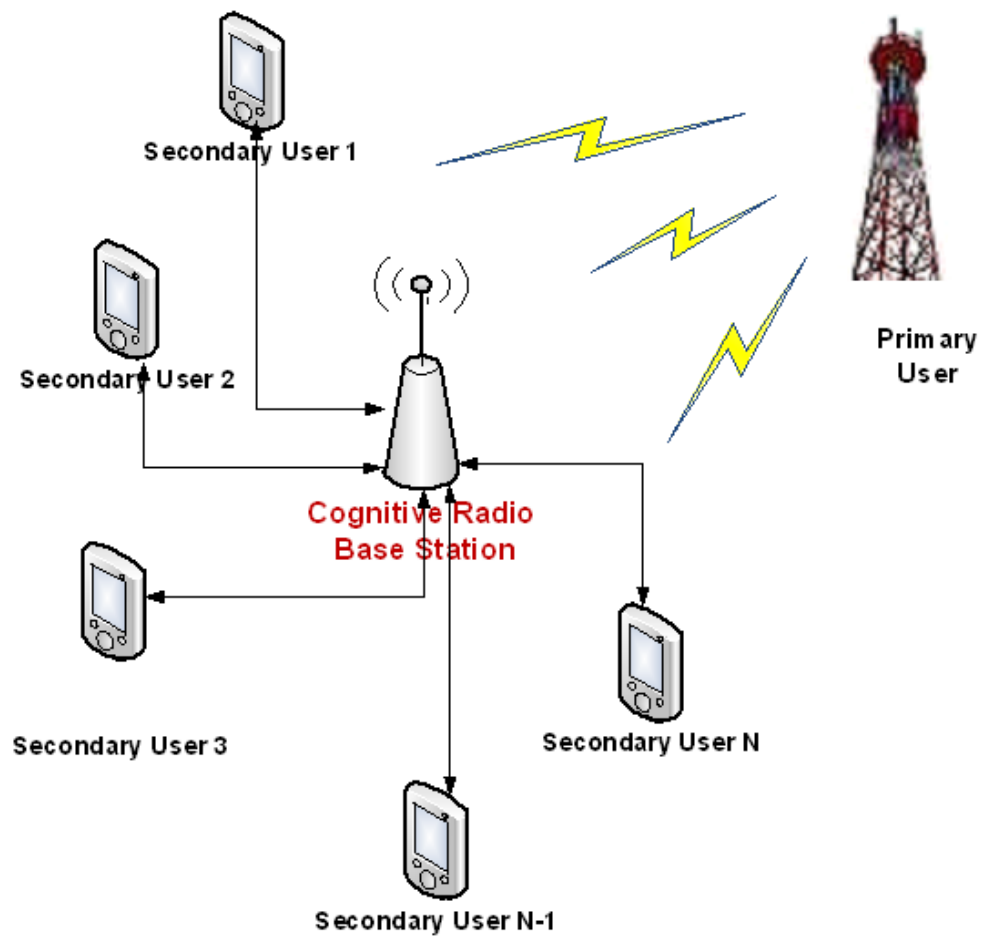

**Fig. 1 Cognitive Radio System**

The state-of-the-art survey of cooperative sensing is presented in [4] to address the issues of cooperation method, cooperative gain, and cooperation overhead. There are different issues and challenges in implementations and designing the cognitive radio are sensitivity, linearity, and dynamic range of RF front end [5]. The detection performance of cooperative spectrum sensing is studied and derived for data fusion and decision fusion over Rayleigh and Raleigh- lognormal fading channels in [6], and wavelet based spectrum sensing is presented and investigated in [10].

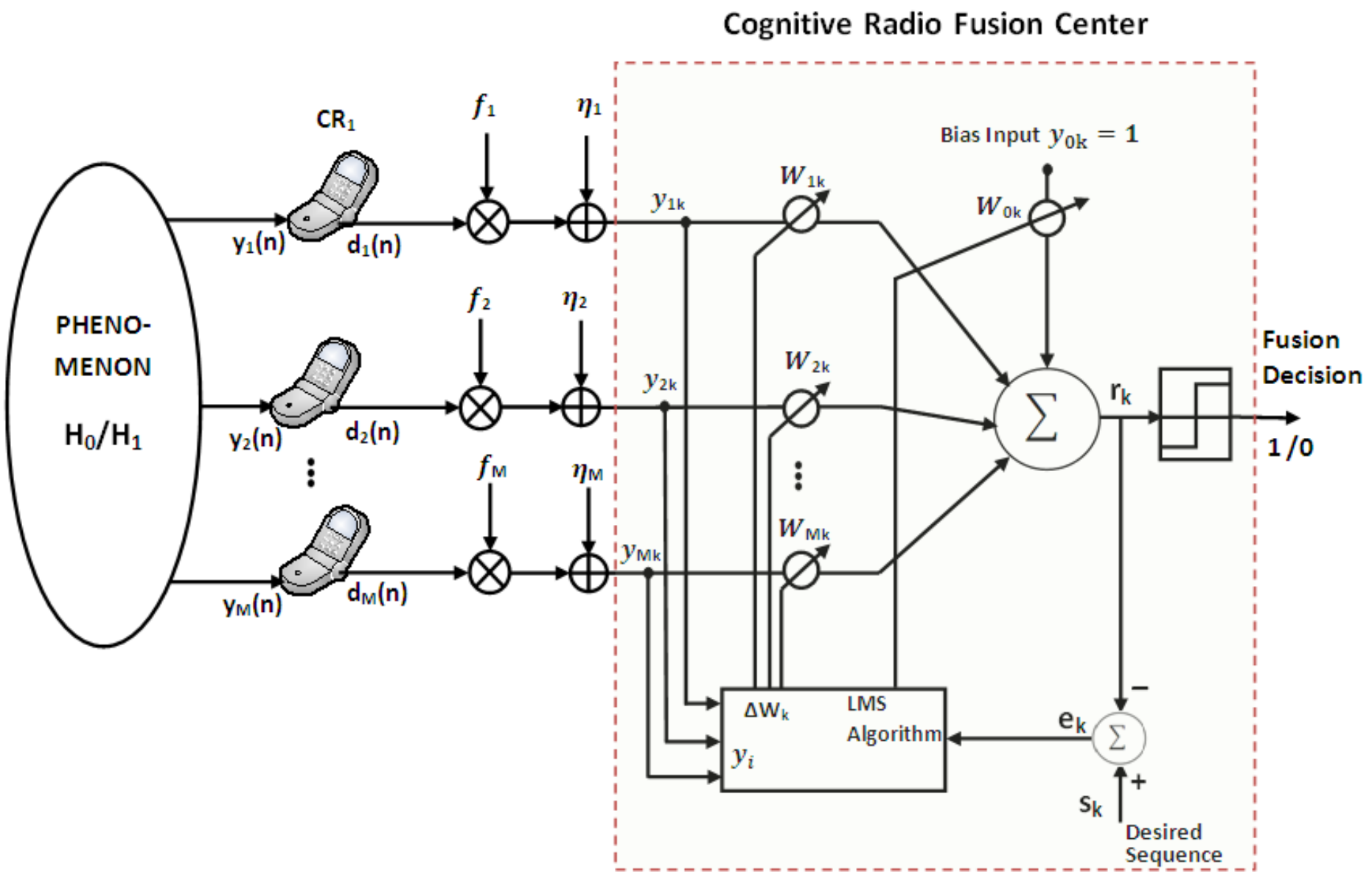

Fig. 2 Adaptive Neural network cooperative spectrum sensing

The spectrum sensing is pivotal for the effective spectrum sharing in cognitive radio. This paper investigates the cognitive radio system using adaptive neural network based data fusion technique for cooperative spectrum sensing. The channel between the secondary user and cognitive radio fusion center or Cognitive radio base station is time varying fading channel, and results severe interference in information and decision from the secondary users. The adaptive spectrum sensing is the most efficient technique for sensing the spectrum intelligently without interfering the primary user. The energy detection technique is used by each secondary user for the spectrum sensing which is simple and efficient.

The remainder of this paper is organized as follows. Section II discusses the system model used for the cooperative spectrum sensing. The proposed neural cooperative spectrum sensing is presented in section III. Performance of proposed neural cooperative spectrum sensing is analyzed and results are presented in section IV. Finally, the section V concludes the paper.

## II. System Model

The cooperative spectrum sensing based on neural network consists of secondary users and to detect the occupancy of the primary, the secondary is with signal processing capabilities should process the primary signal as shown in Fig. 2. The signal at each of the secondary user is modeled as in [12]

$H_0$: $y_i(k) = n_i(k), \;\; i = 1,2,3 \; \ldots.., M$

$H_1$: $y_i(k) = h_i d_i(k) + n_i(k), i = 1,2,3 \; \ldots.., M$ (1)

The received signal is modeled at k$^{th}$ time instant as binary hypothesis, $d_i(k)$ is the primary signal at i$^{th}$ secondary user, and $w_i(k)$ is the complex additive white Gaussian noise [7,8,12] with zero mean and variance $\sigma_i{}^2$, i.e., $n_i(k) \sim \mathcal{CN}(0, \sigma_i{}^2)$. Without loss of generality, $x_i(k)$ and $n_i(k)$ are assumed to be independent of each other. $h_i$ is the gain of the channel between the primary user and the $i^{th}$ secondary user. $H_0$ denotes the primary user is absent, and $H_1$ denotes the primary user is present.

## III. Adaptive Neural Cooperative Spectrum Sensing

The every secondary user in a cognitive radio network is continuously seeking the opportunity for spectrum utilization. The received signal at each secondary user is passed through an ideal bandpass filter with bandwidth B, and by energy detection technique, the energy is computed over an interval of N samples is [12]

$$\boldsymbol{p_i = \sum_{n=1}^{N} |y_i(n)|^2 , i = 1,2,3, \ldots \ldots, M} \quad (2)$$

Where N=2BT is time bandwidth product.

From the Fig.2, the energy of the received signal at secondary user $\{p_i\}$ or the decision at each secondary users $\{CR\}_1^M$ , is reported to the fusion center through reporting channels. The reporting channels are modeled as a multipath fading channel, and the statistics or decision at the fusion center is [12]

$$y_i = f_i d_i + \eta_i, \quad i = 1,2, \ldots \ldots, M \qquad (3)$$

Where $d_i = p_i$ denotes the test statistics is transmitted to the fusion center in the case of soft fusion and $d_i = 0|H_0$ or $d_i = 1|H_1$ when the decision is transmitted to the fusion center in the case of hard fusion. The $f_i$ are i.i.d. $\mathcal{CN}(0, \sigma_{fi}{}^2)$ multipath faded channel gains, $\eta_i$ is white Gaussian noise with zero mean and variance $\sigma_{\eta i}{}^2$, i.e., $\eta_i(n) \sim \mathcal{N}(0, \sigma_{\eta i}{}^2)$.

The false alarm and missed detection probabilities are tradeoffs between spectrum efficiency and interference to primary. The probability of the detection in terms of probability of the false alarm is written as in [12]

$$P_d{}^i = Q\left(\frac{Q^{-1}(P_f{}^i)\sigma_i{}^2\sqrt{2N} + N\sigma_i{}^2}{\sigma_i\sqrt{2N(1+2SNR)}}\right) \quad (4)$$

The lower the false alarm probability leads higher the spectrum utilization and larger the probability of detection (or lower the probability of miss detection, $P_m{}^i = 1 - P_d{}^i$) leads less interference to primary users [12].

$$W_{(K+1)} = W_K + \alpha\frac{e_k Y_k}{|Y_k|^2} \quad (5)$$

and the present linear error is computed as

$$e_k \triangleq d_k - W_k^T X_k \quad (6)$$

The LMS algorithm is self normalizing in the sense that the choice of α does not depend upon [9] magnitude of $Y_k$. The decision about the existence of the primary user is taken based on the result of the thresholding device 1/0. The decision is obtained by applying $r_k$ to the thresholding device, which enables the cognitive radios to adjust their parameters adaptively.

## IV. Results Discussion

The adaptive cooperative cognitive radio system is implemented in MATLAB. The channel between secondary users or cognitive radios and cognitive radio fusion center in the cognitive radio network is assumed nonstationary, and the Rayleigh fading channel model is used for simulation. A maximum Doppler shift of 100 Hz, sample time is of $1 \times 10^{-5}$ Sec , a carrier frequency of 2 GHz and 100 MHz bandwidth are considered for the simulations. The 4-QAM modulated techniques for each secondary user for the communication among the secondary user are used. The decision taken by each cognitive radio is transmitted to the cognitive radio base station via reporting channel, which is bandlimited. The energy statistic is transmitted to the cognitive radio base station is Data fusion and which require high bandwidth. For the neural network architecture 10 numbers of hidden layers are used.

Figure 3 shows the performance of each of the training, validation and test sets. As the network was trained, the mean squared error (mse) decreases rapidly, and best validation performance is found at $3.5531 \times 10^{-7}$. Figure 4 shows the gradient and validation check, the gradient is about $8.3542 \times 10^{-7}$and the validation check is about 0 at epoch 27 respectively. The weight and bias is adjusted to Optimum value which is shown as a low value of gradient and better network learning which results network more accurate and reliable, reduced false predictions. The sufficient network learning shown in validation plot, and the stop point of training and the starting of the over fitting of the data. Table 1 gives the mean square error and percentage of error during the training, validation and testing, which shows the proper classification of samples.

TABLE 1 MEAN SQUARE ERROR AND % ERROR

| | Samples | MSE | % Error |
|---|---|---|---|
| **Training** | 2 | 3.55307e-7 | 2425 |
| **Validation** | 0 | 4.94684e-1 | 50 |
| **Testing** | 0 | 4.94684e-1 | 50 |

Figure 5 shows the how error sizes are distributed at different instances as an error histogram, and the error approaches to zero. Figure 6 shows the neural network receiver operating characteristic for data and relation between the false positive and true positive rates as the output thresholding is varied from 0 to 1. The training, validation, test and collective ROC describe the neural network performance.

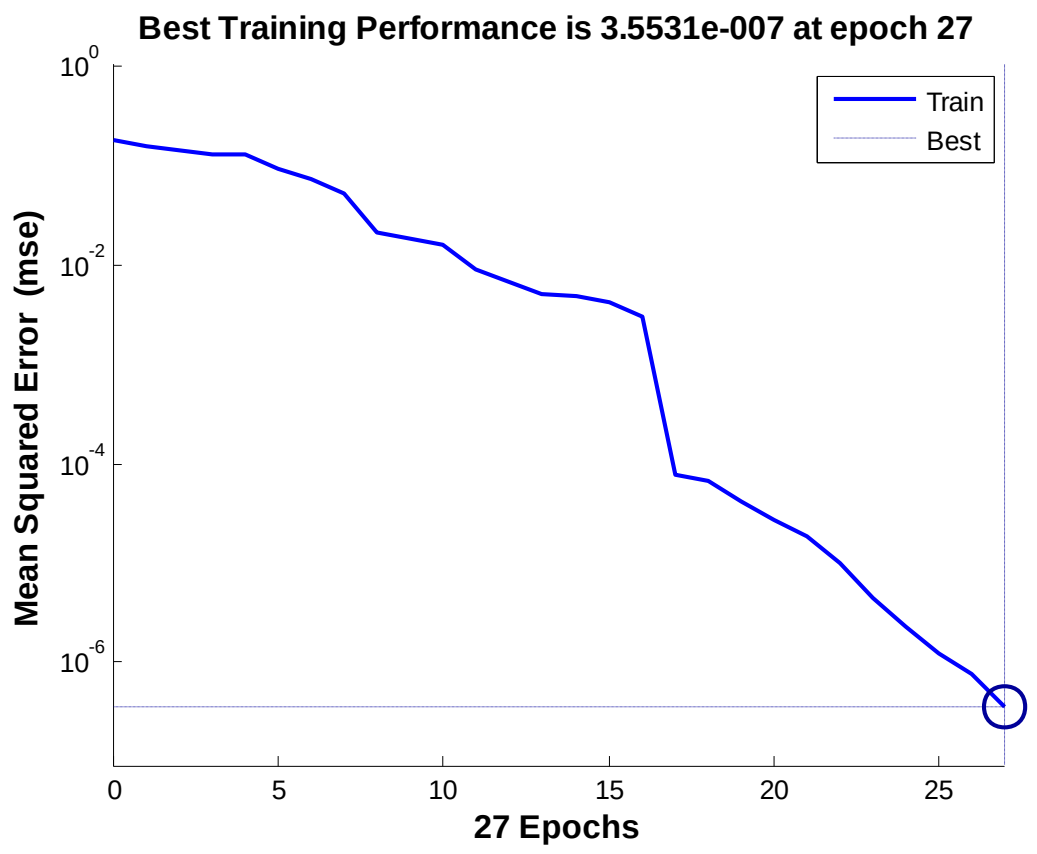

Fig. 3 Mean square Error performance

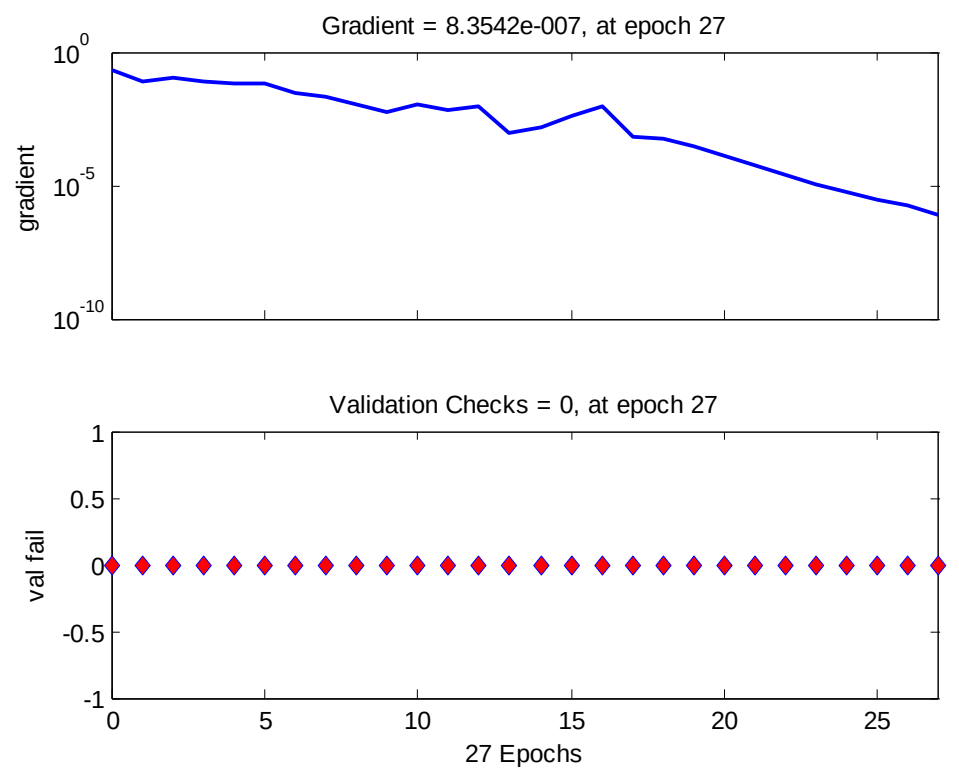

Fig. 4 The gradient and validation check

Figure 7 shows the confusion matrix is across all samples. The confusion matrix shows the percentages of correct and incorrect classifications.The designed network performance in the large number of samples is evaluated and the correct classifications are shown in green squares and incorrect classifications in the red squares.

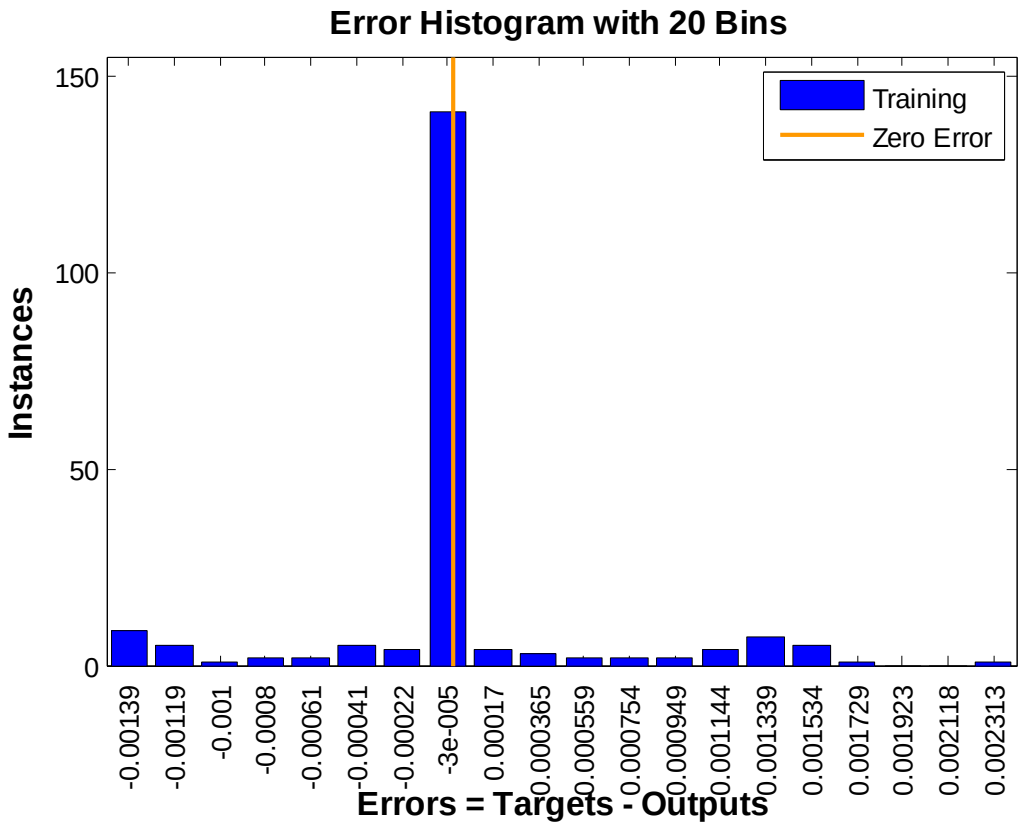

Fig. 5 Error histogram of CR System

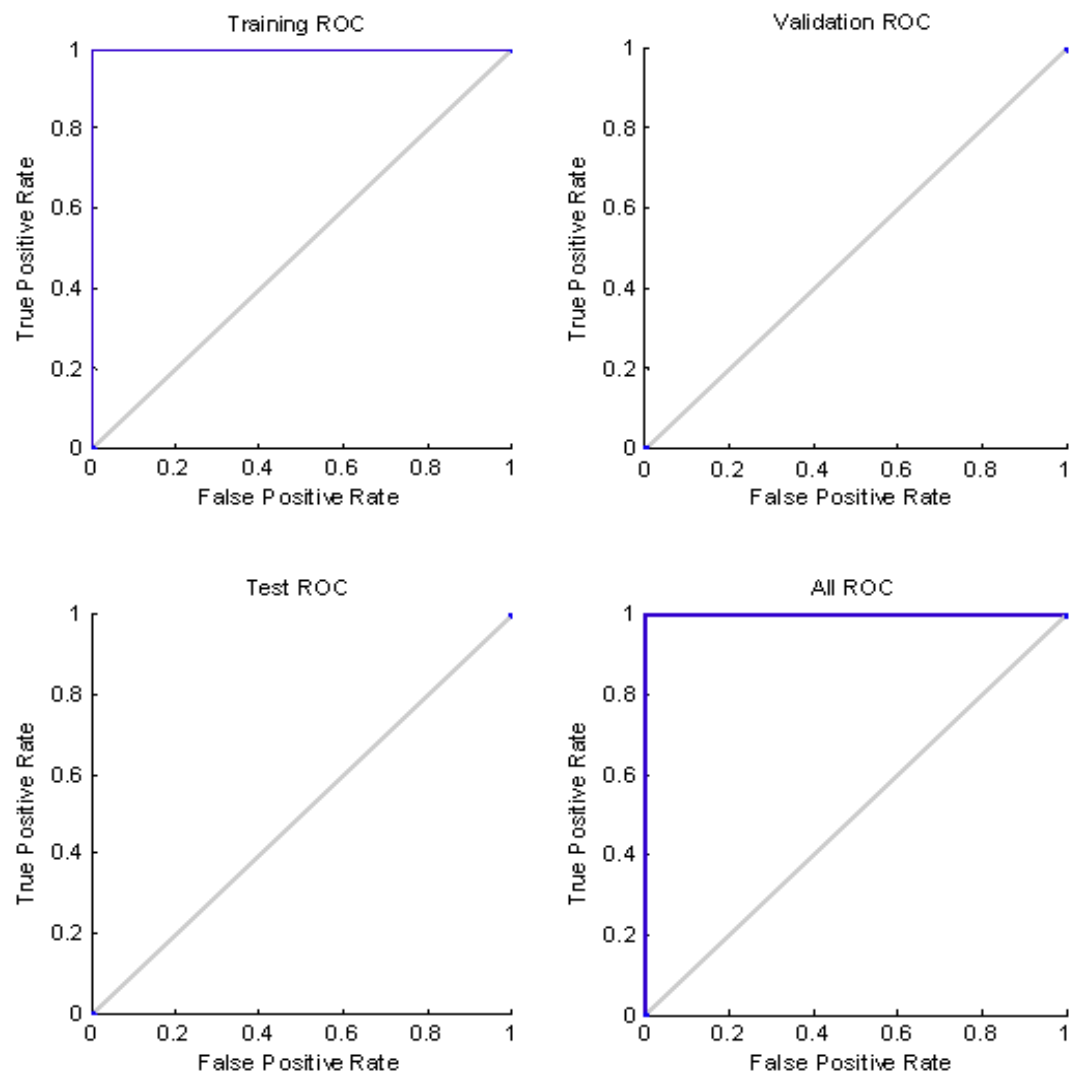

Fig. 6 Receiver operating characteristic of Neural Network

Figure 8 shows the ROC curve of the behavior of the system in the multipath fading and shadowing channel environment with adaptive neural network and majority rule. The neural network adaptive system performance is significantly improved over the majority rule. The less complexity in the proposed system is observed in the design and implementation, and it outperforms over majority rule but the convergence time of proposed system may increase as a number of the secondary users in the cognitive radio system

## V. Conclusion

In this paper, the time varying channel among the cognitive radios and the reporting channel between the cognitive radio and cognitive radio base station considered. Due to multipath fading and shadowing, the sensing performance deteriorated and the probability of false alarm and probability of miss detection increased. The sensing performance may improve if the all the secondary users have exactly detected the existence of the primary user in the band of interest.

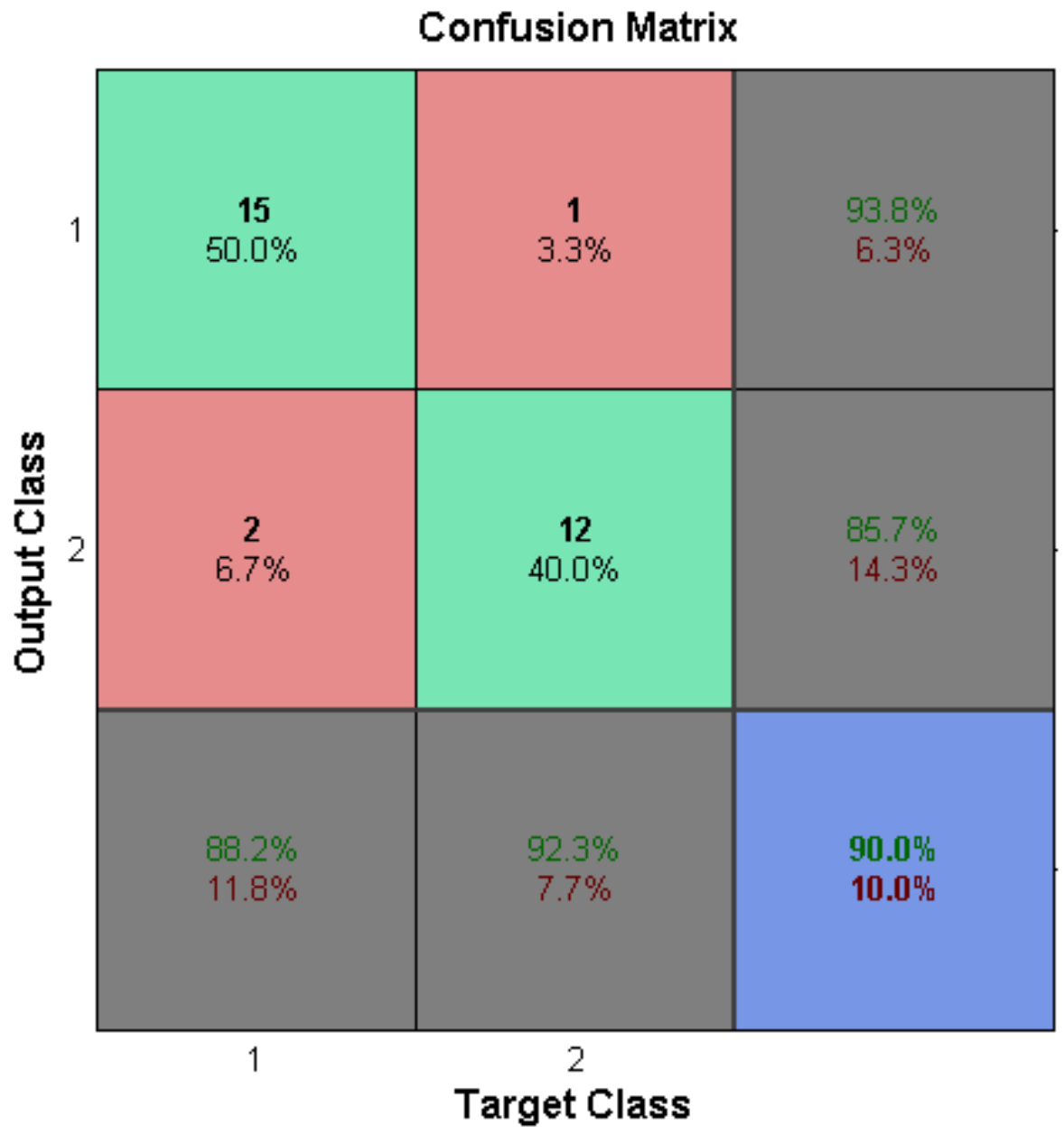

Fig. 7 Confusion matrix of CR system

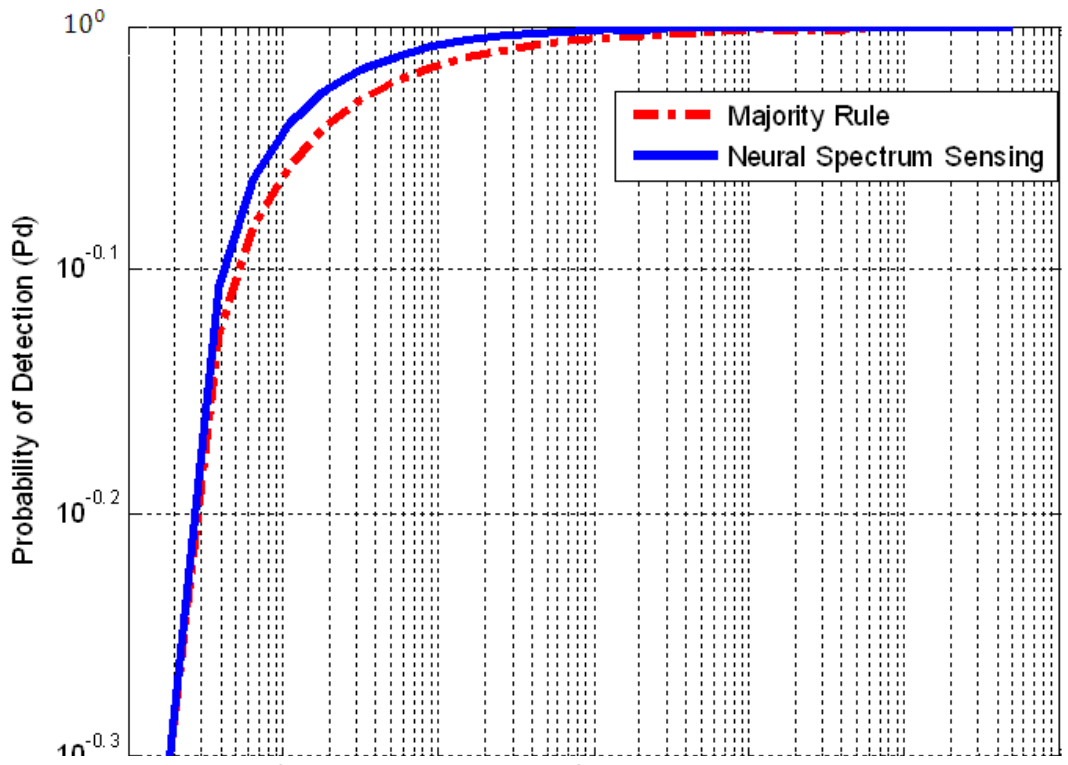

Fig. 8 The performance curve of neural adaptive CR system.

The adaptive neural network spectrum sensing technique proposed and investigated the performance over the AND, OR and Majority rule of decision fusion. The performance of the proposed method improved significantly over the majority rule and the proposed method outperform the time varying multipath fading channel environment.

In order to reduce the circuit and signal processing complexity at the secondary user information fusion could be the best choice. The proposed method is more efficient in the case of information fusion at cognitive radio base station as compared to the other complex methods of information fusion. The FPGA implementation of the proposed method has less complexity and the detection performance is improved significantly is in worst channel conditions with decision fusion and information fusion.

## References

[1] T. Yucek and H. Arslan, "A survey of spectrum sensing algorithms for cognitive radio applications," *IEEE Communications Surveys & Tutorials*, vol. 11, no. 1, pp. 116-130, Mar. 2009.

[2] Federal Communications Commission, "Spectrum Policy Task Force," Rep. ET Docket no. 02-135, Nov. 2002.

[3] S. Haykin, "Cognitive radio: brain-empowered wireless communications," *Selected Areas in Communications, IEEE Journal on*, vol. 23, no. 2, pp. 201-220, Feb. 2005.

[4] Ian F. Akyildiz, Brandon F. Lo, Ravikumar Balakrishnan, Cooperative spectrum sensing in cognitive radio networks: A survey, Physical Communication, Volume 4, Issue 1, March 2011, Pages 40-62, ISSN 1874-4907, 10.1016/j.phycom.2010.12.003.

[5] Cabric, D.; Mishra, S.M.; Brodersen, R.W.; , "Implementation issues in spectrum sensing for cognitive radios," *Signals, Systems and Computers, 2004. Conference Record of the Thirty-Eighth Asilomar Conference on* , vol.1, no., pp. 772- 776 Vol.1, 7-10 Nov. 2004

[6] Atapattu, S.; Tellambura, C.; Hai Jiang; , "Energy Detection Based Cooperative Spectrum Sensing in Cognitive Radio Networks," *Wireless Communications, IEEE Transactions on* , vol.10, no.4, pp.1232-1241, April 2011.

[7] M. Barkat, Signal Detection and Estimation (Artech House Radar Library), 2nd ed. Artech House Publishers, Jul. 2005.

[8] D. Tse and P. Viswanath, *Fundamentals of Wireless Communication*. Cambridge University Press, Jul. 2005.

[9] Widrow, B.; Lehr, M.A.; , "30 years of adaptive neural networks: perceptron, Madaline, and backpropagation," *Proceedings of the IEEE* , vol.78, no.9, pp.1415-1442, Sep 1990 doi: 10.1109/5.58323

[10] D.S. Aldar,"Centralized Integrated Spectrum Sensing for Cognitive Radios" *International Journal of Computer Science & Communication*, Vol. 1, No. 2, pp. 277-279, July-December 2010.

[11] Dilip S. Aldar, B. B. Godbole," Performance analysis of neural network based time domain adaptive equalization for OFDM", *International Journal of Recent Trends in Engineering*, Vol 2, No. 4, November 2009.

[12] Dilip S. Aldar, "Distributed Fuzzy Optimal Spectrum Sensing In Cognitive Radio", *Progress In Electromagnetics Research Journal*, 2012. [Under Review]

[13] Satish Kumar, Neural Network: A Classroom Approach. Tata McGraw-Hill Publication, Jun 2004.

## Authors' information

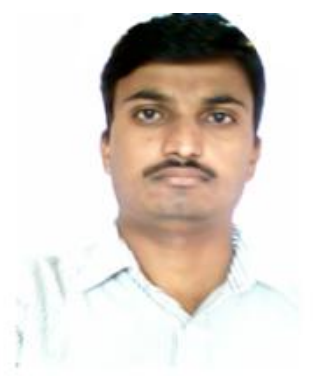

**Dilip S. Aldar** born in India received the B.Eng. degree from Shivaji University, India, in 2001, M.Tech from Dr. BATU, Lonere, India. He is currently working toward the Ph.D. degree in the area wireless communication.

He is currently Assistant professor in Electronics Engineering, K B P College of Engineering & polytechnic, Satara, India. He is working on Dynamic Spectrum Management and transmits power control algorithms for cognitive radio.

His area of research includes game theoretic approach for cognitive radio network, Genetic algorithms for cognitive radio.
.